\documentclass[sigconf]{acmart}

\usepackage{mathtools}
\usepackage{algorithm}
\usepackage{algorithmic}
\usepackage{multirow}
\usepackage{makecell}
\usepackage{colortbl}

\AtBeginDocument{%
  }

\copyrightyear{2026}
\acmYear{2026}
\setcopyright{cc}
\setcctype{by-nc-nd}
\acmConference[MM '26]
  {Proceedings of the 34th ACM International Conference on Multimedia}
  {November 10--14, 2026}
  {Rio de Janeiro, Brazil}
\acmBooktitle{Proceedings of the 34th ACM International Conference on Multimedia (MM '26), November 10--14, 2026, Rio de Janeiro, Brazil}
\acmDOI{10.1145/3767308.3836335}
\acmISBN{979-8-4007-2213-4/2026/11}

\begin{document}

\title[RADIANT for Universal Fusion]{RADIANT: Redundancy-Aware Information Allocation for Universal Multi-Modal Fusion}
\author{Shenghai Wu}
\authornote{Corresponding author.}
\orcid{0009-0001-9473-136X}
\affiliation{%
  \department{Research Institute}
  \institution{Newland Digital Technology}
  \city{Fuzhou}
  \country{China}}
\email{stevenwsh01@gmail.com}

\author{Yue Lin}
\orcid{0009-0008-7758-4530}
\affiliation{%
  \department{Research Institute}
  \institution{Newland Digital Technology}
  \city{Fuzhou}
  \country{China}}
\email{yuelin109@gmail.com}

\renewcommand{\shortauthors}{Shenghai Wu and Yue Lin}

\begin{abstract}
Universal multi-modal fusion often fails not because modalities are insufficiently aligned, but because the fusion process allocates information poorly: redundant evidence is repeatedly amplified, modality-unique evidence is over-smoothed during alignment, and task-dependent synergistic evidence is insufficiently activated. We present \textbf{RADIANT} (Redundancy-Aware Dynamic Information Allocation Network), a task-general framework that organizes fusion around three complementary roles: suppressing redundant cross-modal overlap, preserving modality-unique evidence, and composing task-conditioned synergy through cross-modal interaction. Specifically, stable matrix-exponential projections are used to reduce interference among shared and modality-unique streams, while task memory with hard positive updates selects which synergistic interactions should be activated for the current fusion regime. Extensive experiments across six heterogeneous fusion tasks show consistent improvements over the compared methods. On FLIR, RADIANT improves downstream detection mAP@0.5 from 91.2\% to 96.8\%, a gain of 5.6 percentage points (6.1\% relative) over GIF. RADIANT also exhibits strong direct cross-dataset transfer, with further gains from inference-time task-memory adaptation while keeping all network weights frozen. Additional ablations and sensitivity studies support the robustness of the allocation mechanism. Code: \url{https://github.com/StevenWSH27/RADIANT}.
\end{abstract}


\maketitle

\section{Introduction}

\begin{figure}[t]
\centering
\includegraphics[width=0.98\linewidth]{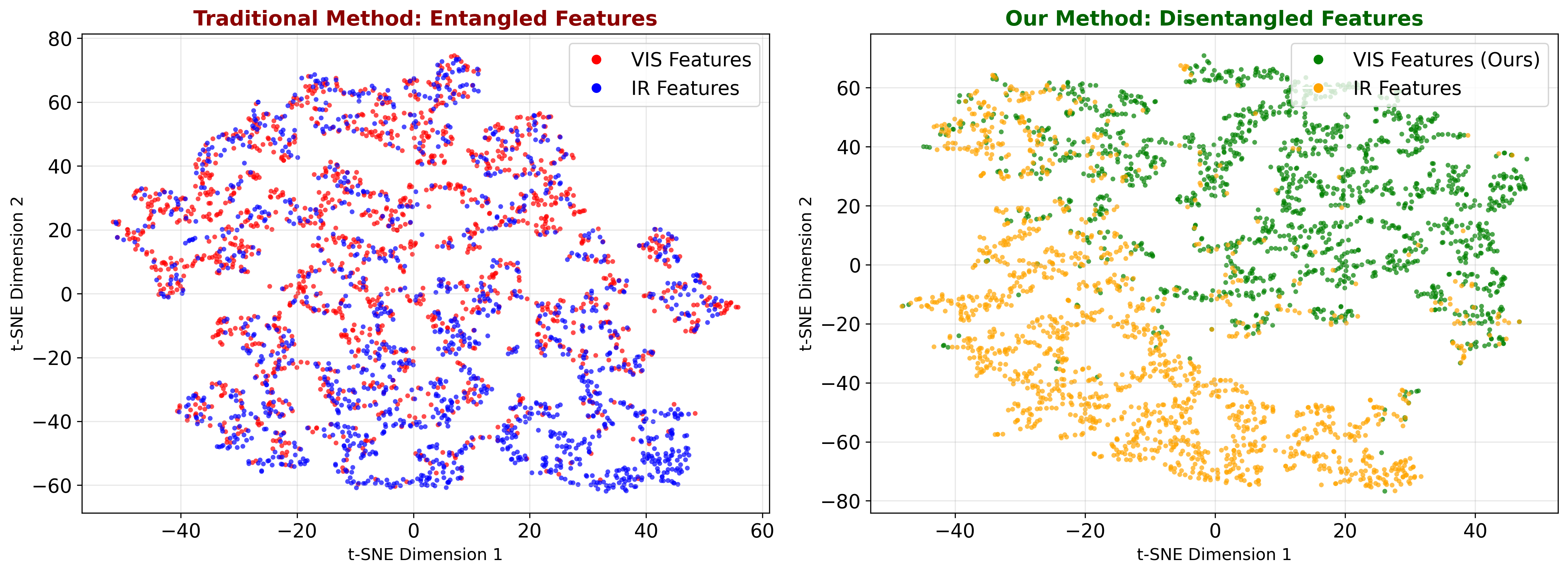}
\caption{Information allocation failure versus allocation-aware fusion. \textbf{Left:} Conventional fusion over-emphasizes redundant cross-modal consistency and suppresses informative modality differences, causing ambiguous representations. \textbf{Right:} RADIANT separates non-interfering shared evidence ($\mathbf{F}_r$), modality-unique evidence ($\mathbf{F}_{u1}, \mathbf{F}_{u2}$), and task-conditioned synergistic evidence ($\mathbf{F}_s$) for controllable universal fusion.}
\Description{
Conceptual comparison between conventional multimodal fusion and the
proposed allocation-aware fusion strategy. The left side illustrates
conventional fusion, where strongly overlapping information from two
modalities is repeatedly emphasized while modality-specific information
is weakened, producing an entangled fused representation. The right side
illustrates RADIANT, which organizes information into a shared stream
representing cross-modal common evidence, two modality-unique streams
retaining information specific to each input modality, and a
task-conditioned synergistic stream formed through cross-modal
interaction. These complementary streams are then combined to construct
the fused representation.
}
\label{fig:feature_entanglement}
\end{figure}

Multi-modal image fusion is central to medical imaging, autonomous
systems, surveillance, and remote sensing, where performance depends on combining complementary sensor evidence without destroying task-relevant details. A central challenge in universal multimodal fusion is not only feature alignment, but also the allocation of redundant, modality-unique, and interaction-dependent information across tasks. In practice, fusion quality can deteriorate when redundant evidence is repeatedly reinforced, modality-unique evidence is washed out during alignment, and synergistic evidence that emerges through cross-modal interaction is under-modeled~\citep{wu2024muse, cicchetti2025gramian,
dufumier_castillo2025}.

Current methods still fall into two broad families: (1) \emph{task-specific architectures}~\citep{ref52, ref53, ref41} that optimize for particular modality pairs but transfer poorly, and (2) \emph{generalized frameworks}~\citep{ref38, ref5} that broaden applicability yet often rely on a single shared fusion space. The former struggle to scale across application regimes, while the latter frequently over-index on cross-modal consistency. As a result, they often preserve what both modalities already know, but fail to decide what should be suppressed, what must be retained independently, and what should be created only after interaction.

This perspective suggests a different formulation of universal fusion. Rather than treating the problem as generic feature disentanglement, we cast it as \emph{task-conditioned information allocation}: allocate redundant/shared evidence conservatively, preserve modality-unique evidence explicitly, and compose synergistic evidence through task-aware interaction. Under this view, orthogonality is useful only insofar as it provides a stable non-interference bias, while task memory is valuable because it selects the interaction patterns that are beneficial for the current fusion regime.

Based on this principle, we present \textbf{RADIANT}, a redundancy-aware information allocation framework for universal multi-modal fusion. RADIANT combines a lightweight hybrid Vision Mamba--Swin backbone with a stable non-interfering shared/unique decomposition and task-memory-guided cross-modal interaction. Its contributions are:

\begin{enumerate}
\item \textbf{Task-conditioned information allocation for universal fusion.} We introduce a fusion framework that explicitly distinguishes redundant shared evidence, modality-unique evidence, and task-conditioned synergistic evidence, allowing a single model to adapt its information allocation across six heterogeneous fusion tasks.

\item \textbf{Synergy-aware fusion with task memory.} We reinterpret prototype learning as task memory that routes and stabilizes cross-modal interactions. Hard positive updates improve intra-task compactness, enabling the model to activate the synergistic interactions most useful for each fusion regime without task-specific architectural redesign.

\item \textbf{Lightweight universal performance with robust transfer.}
With a model footprint of only 3.108 MB and 9.42 GFLOPs, RADIANT achieves state-of-the-art results across six fusion tasks and shows robust direct cross-dataset transfer, with further improvements through inference-time task-memory adaptation while keeping network weights frozen.
\end{enumerate}

Across six fusion tasks, RADIANT consistently outperforms the compared methods. On FLIR, it improves downstream detection mAP@0.5 from 91.2\% to 96.8\% over GIF, corresponding to a gain of 5.6 percentage points (6.1\% relative). Systematic hyperparameter perturbations produce at most 1.2\% median absolute percentage deviation within the evaluated ranges.

\section{Related Work}

\subsection{Universal and Task-Specific Multi-Modal Fusion}

Multi-modal image fusion has progressed from handcrafted priors~\citep{ref13} to deep architectures spanning infrared-visible, multi-focus, medical, and remote-sensing settings. Task-specific methods such as CDDFuse~\citep{ref52}, DDFM~\citep{ref53}, Text-IF~\citep{ref41}, CoCoNet~\citep{ref23}, P2Sharpen~\citep{ref47}, ZeroSharpen~\citep{ref34}, and RTFusion~\citep{RTFusion2023} achieve strong in-domain performance by specializing for a narrow fusion regime, but transfer poorly across tasks. Universal methods such as U2Fusion~\citep{ref38}, MUFusion~\citep{ref5}, SDNet~\citep{ref43}, UNIFusion~\citep{ref4}, UniFuse~\citep{UniFuse2025}, and MdaIF~\citep{MdaIF2025} broaden applicability, yet often depend on a single latent space or a fixed fusion rule. Recent task-general systems such as GRAM~\citep{cicchetti2025gramian}, CoMM~\citep{dufumier_castillo2025}, AdaSFFuse~\citep{adasffuse2025}, and C2RF~\citep{C2RF2025} move toward more principled multimodal representations, but they still provide limited control over how redundant, unique, and interaction-derived evidence are allocated during fusion, especially under changing task regimes.

\subsection{Information Allocation in Multi-Modal Representation Learning}

A large body of multimodal work separates shared and modality-specific factors~\citep{tsai2019mfm, zhao2023s2fnet}. Related domain-translation and fusion models further preserve domain- or modality-specific characteristics~\citep{ren2018fan, hu2022ssgan}. DecAlign~\citep{decalign2025} further decouples common and unique representations through hierarchical alignment, while related orthogonalized or de-redundancy approaches improve non-interference between streams. These methods show the utility of structured factorization, but most stop at common/private decomposition. They do not directly address the 2025--2026 shift toward asking which information should be shared, which should be preserved independently, and which should be synthesized only through cross-modal interaction. Recent multimodal studies increasingly frame this question through redundancy, uniqueness and synergy. CoMM~\citep{dufumier_castillo2025} explicitly uses information decomposition ideas to reason about complementary features, and newer multimodal systems explore orthogonality, de-redundancy, or expert routing as means to better distribute overlapping versus distinctive information. Our work follows this direction in the fusion setting: orthogonality is treated as an implementation choice for non-interference, while the main objective is allocation of redundant, unique, and task-conditioned synergistic evidence.

\subsection{Task Memory and Cross-Modal Interaction}

Contrastive learning has become widely used for cross-modal alignment and multimodal representation learning~\citep{ascl2024, zhang2024multimodal, zhang2024multigrained, tang2025fakenews, yuan2026crossmodal, yuan2026moe}. Pixel-, region-, and prototype-based objectives provide different levels of cross-modal supervision. Task-aware representation learning is also closely related to multi-task learning, where shared representations can improve generalization across related tasks~\citep{ref20}. Recent systems such as Memory Aware Synapses~\citep{MemoryAwareSynapses2025}, multi-query cross-modal attention fusion~\citep{MultiQueryCrossModalAttention2025}, and FMFR~\citep{FMFR2026} further suggest that adaptive memory or query-conditioned interaction is useful when downstream regimes differ. Building on this line of work, RADIANT uses task prototypes as \emph{task memory} to determine which cross-modal interactions should be emphasized, enabling adaptive synergy composition across heterogeneous fusion tasks.

\section{Methodology}

\begin{figure*}[t]
\centering
\includegraphics[width=0.95\textwidth]{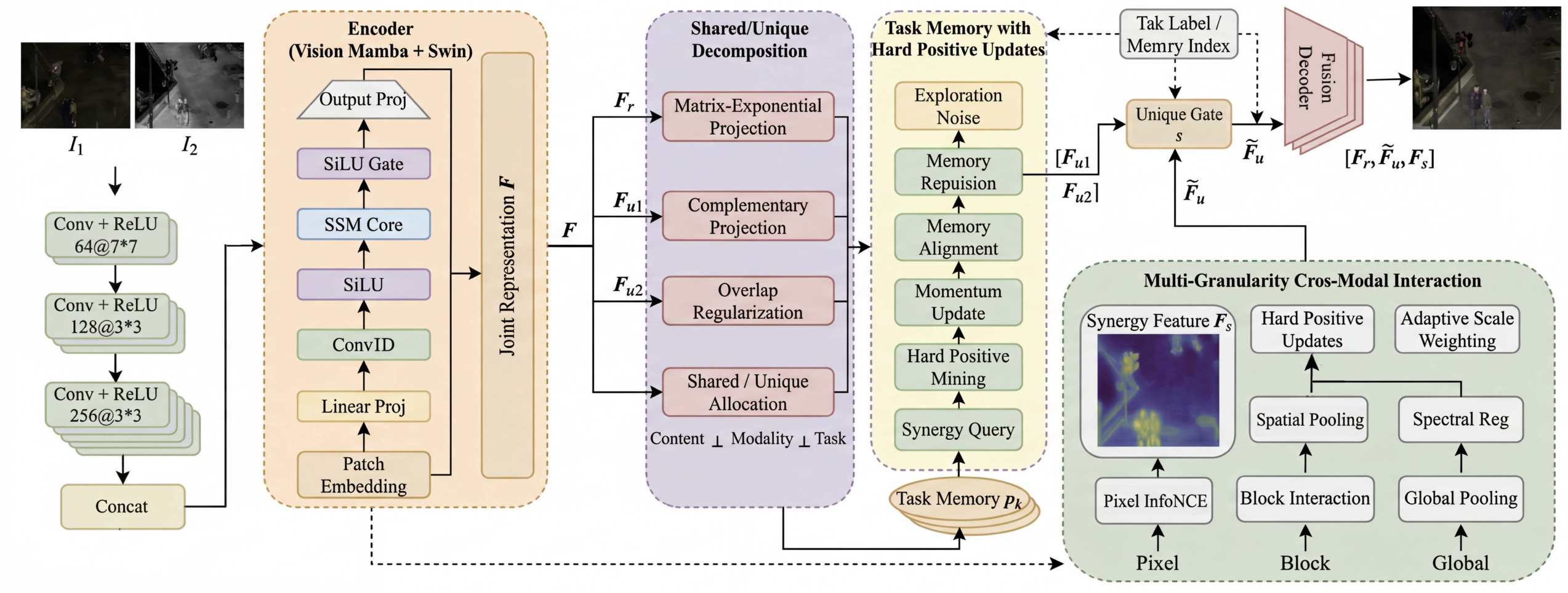}
\caption{RADIANT architecture overview. A dual-branch Vision Mamba--Swin encoder first maps the two modalities to a joint representation. A shared/unique decomposition module then separates shared and modality-unique evidence ($\mathbf{F}_r$, $\mathbf{F}_{u1}$, $\mathbf{F}_{u2}$), and task memory guides multi-granularity interaction to compose synergistic evidence ($\mathbf{F}_s$) before decoding.}
\Description{
Overview of the RADIANT architecture from two input modalities to the
final fused image. The two source images are first processed by
modality-specific convolutional branches and combined into a joint
representation using a lightweight Vision Mamba--Swin encoder. The joint
features are passed to a shared/unique decomposition module containing
matrix-exponential projection, complementary projection, and overlap
regularization, producing one shared feature stream and two
modality-aware unique streams. In parallel, a task-memory module forms a
synergy query, performs hard-positive mining, updates task-memory anchors
with momentum, and aligns the current representation with the task
memory. A multi-granularity cross-modal interaction module combines
pixel-level, block-level, and global interactions with adaptive
granularity weighting to produce the synergistic feature stream. A
task-memory-conditioned gate selects useful unique information, and the
decoder combines the shared, gated unique, and synergistic streams to
generate the final fused image.
}
\label{fig:architecture}
\end{figure*}

\subsection{Problem Formulation and Information Allocation Principle}

Let $\mathbf{I}_1, \mathbf{I}_2 \in \mathbb{R}^{H \times W \times C}$ denote source images from two modalities. Encoder $\phi_\theta$ maps them to a joint representation $\mathbf{h} \in \mathbb{R}^{N \times D}$, and decoder $\psi_\omega$ predicts the fused output $\mathbf{Y}$. For $K$ fusion tasks, training data are organized as $\mathcal{D}_k = \{(\mathbf{I}_1^{(n)}, \mathbf{I}_2^{(n)})\}_{n=1}^{N_k}$.

Instead of describing universal fusion as generic disentanglement, RADIANT models it as \emph{information allocation}. The representation is organized into three roles: redundant/shared evidence $\mathbf{F}_r$, modality-unique evidence $\mathbf{F}_{u1}, \mathbf{F}_{u2}$, and synergistic evidence $\mathbf{F}_s$ that emerges only after task-conditioned cross-modal interaction. A desirable fusion model should suppress unnecessary redundancy, preserve modality-unique cues, and activate synergy only when the current task regime benefits from it. We express this principle as
\begin{equation}
\min_{\phi_\theta,\psi_\omega} \; \mathcal{R}_{\text{red}}(\mathbf{F}_r) + \lambda_u\,\mathcal{R}_{\text{uniq}}(\mathbf{F}_{u1},\mathbf{F}_{u2}) - \lambda_s\,\mathcal{G}_{\text{syn}}(\mathbf{F}_s; \mathbf{Y})
\label{eq:allocation_principle}
\end{equation}
where $\mathcal{R}_{\text{red}}$ penalizes redundant amplification, $\mathcal{R}_{\text{uniq}}$ preserves the modality-specific evidence, and $\mathcal{G}_{\text{syn}}$ rewards the task-useful synergy. In practice these quantities are instantiated through fusion, non-interference, and task-memory-guided interaction losses rather than by exact information decomposition.

The overall architecture in Figure~\ref{fig:architecture} realizes this principle in two stages. First, a stable non-interfering parameterization separates shared and modality-unique streams. Second, task memory and multi-granularity cross-modal interaction compose synergistic evidence conditioned on the active fusion regime. The orthogonality-inducing parameterization provides a non-interference bias for the shared and modality-unique streams, while task memory governs task-conditioned synergy composition.

We interpret the learned streams as an approximate functional factorization. When a single latent space jointly represents shared evidence, modality-specific cues, and interaction-dependent effects, optimization may over-share information and suppress weak but task-critical distinctive cues. The sequential shared/unique decomposition and task-memory-guided interaction provide a structured training bias that mitigates this interference while retaining task-dependent flexibility.

Under this interpretation, RADIANT should exhibit three practical behaviors that we later verify empirically: reduced harmful overlap between shared and non-shared streams, better preservation of modality-specific cues useful for reconstruction, and stronger task-dependent interaction gains when the best fusion policy varies across datasets. Supplementary Secs.~A and S2 provide the extended discussion and diagnostics.

This viewpoint also clarifies why a universal model can remain lightweight. RADIANT does not attempt to learn a separate backbone or decoder for each fusion regime. Instead, the encoder supplies a common joint representation, while the decomposition and memory-guided interaction stages determine how much information should remain shared, stay modality-specific, or be recombined as synergy for the current task. In other words, universality is achieved by reallocating information within a shared pipeline rather than by multiplying task-specific capacity. This is the reason the model can cover heterogeneous benchmarks without abandoning a compact backbone or resorting to full task-wise retraining.

Our empirical analyses in Supplementary Secs.~A and S2 evaluate whether this allocation strategy reduces harmful overlap while preserving task-relevant signal. The semantic roles of the learned streams are determined jointly by $\mathcal{L}_{\text{fusion}}$, $\mathcal{L}_{\text{align}}$, and $\mathcal{L}_{\text{orthogonal}}$.

\subsection{Non-Interfering Shared/Unique Decomposition}

Given joint features $\mathbf{F} \in \mathbb{R}^{N \times D}$, we construct a shared stream and two modality-aware residual streams using learnable projections
\begin{equation}
\mathbf{F}_r = \mathbf{F}\mathbf{Q}_r, \qquad \mathbf{F}_{u1} = \mathbf{F}\mathbf{Q}_{u1}, \qquad \mathbf{F}_{u2} = \mathbf{F}\mathbf{Q}_{u2},
\end{equation}
where the projection matrices are parameterized sequentially as
\begin{align}
\mathbf{Q}_r &= \exp(\mathbf{A}_r - \mathbf{A}_r^\top)\mathbf{W}_r, \\
\mathbf{Q}_{u1} &= \exp(\mathbf{A}_{u1} - \mathbf{A}_{u1}^\top)(\mathbf{I} - \mathbf{Q}_r\mathbf{Q}_r^\top)\mathbf{W}_{u1}, \\
\mathbf{Q}_{u2} &= \exp(\mathbf{A}_{u2} - \mathbf{A}_{u2}^\top)(\mathbf{I} - \mathbf{Q}_r\mathbf{Q}_r^\top - \mathbf{Q}_{u1}\mathbf{Q}_{u1}^\top)\mathbf{W}_{u2}.
\end{align}
Here $\mathbf{A}_\cdot \in \mathbb{R}^{D \times D}$ are learnable skew generators and $\mathbf{W}_\cdot \in \mathbb{R}^{D \times D/3}$ are projection weights. The matrix exponential contributes an orthogonal preconditioner with condition number 1, which stabilizes the basis rotation before task-adaptive shaping by $\mathbf{W}_\cdot$. We do \emph{not} claim that the full matrices $\mathbf{Q}_r, \mathbf{Q}_{u1}, \mathbf{Q}_{u2}$ remain exactly orthogonal or perfectly conditioned after multiplication with $\mathbf{W}_\cdot$; likewise, the complement operators in Eqs.~(4)--(5) should be interpreted as \emph{approximate non-interference projectors}, not exact geometric projectors.

This approximation is intentional. Using the orthogonal factor alone was empirically less expressive in preliminary experiments, while the additional $\mathbf{W}_\cdot$ terms allow task-adaptive subspace shaping. Stability in our setting therefore comes from the combination of orthogonal preconditioning, sequential complement masking, and explicit overlap regularization, rather than from any claim that the learned subspaces are exactly mutually orthogonal. We enforce near-semi-orthonormality with Eq.~(6) and monitor the resulting overlap rather than asserting an exact decomposition. The regularizer is

\noindent\textbf{Projection sanity check.}
On LLVIP, the overlap penalty in Eq.~(6) drives the learned columns
toward near-semi-orthonormality. Supplementary Sec.~S5 compares the
matrix-exponential parameterization with simpler alternatives and
shows the best accuracy--stability trade-off.
\begin{equation}
\mathcal{L}_{\text{orthogonal}} = \sum_{i \neq j} \|\mathbf{F}_i^\top \mathbf{F}_j\|_F^2 + \lambda_{\text{reg}} \sum_{i \in \{r,u1,u2\}} \|\mathbf{Q}_i^\top \mathbf{Q}_i - \mathbf{I}\|_F^2,
\end{equation}
where $\{\mathbf{F}_i\} = \{\mathbf{F}_r, \mathbf{F}_{u1}, \mathbf{F}_{u2}\}$ and $\lambda_{\text{reg}} = 0.01$. Table~\ref{tab:projection_sanity} summarizes the average orthogonality residual measured on LLVIP and shows that the full parameterization remains close to semi-orthonormal in practice while outperforming the orthogonal-only variant.

\begin{table}[!htbp]
\centering
\scriptsize
\caption{Projection sanity check on LLVIP. Residual is the average $\|\mathbf{Q}_i^\top\mathbf{Q}_i-\mathbf{I}\|_F$ over $i\in\{r,u1,u2\}$ on the validation split. Lower is better.}
\label{tab:projection_sanity}
\begin{tabular}{lcc}
\toprule
Variant & Residual $\downarrow$ & EI $\uparrow$ \\
\midrule
Orthogonal factor only ($\mathbf{W}_\cdot = \mathbf{I}$ / fixed basis) & 0.021 & 65.31 \\
Full sequential parameterization & 0.034 & 66.08 \\
\bottomrule
\end{tabular}
\end{table}

Supplementary Sec.~S5 compares matrix exponential against QR, Cayley, and linear alternatives, showing it is the most stable implementation among the tested choices.

\subsection{Task Memory with Hard Positive Updates}

For each task $k \in \{1, \ldots, K\}$, we maintain a learnable memory anchor $\mathbf{p}_k \in \mathbb{S}^{d-1}$ that summarizes the interaction regime preferred by that task. Rather than treating prototypes as the endpoint of supervision, RADIANT uses them to modulate which synergistic interactions should be activated. For a training sample $i$ from task $k$, the synergy-conditioned stream is first pooled and projected into the memory space,
\begin{equation}
\bar{\mathbf{f}}_{s,k}^{(i)} = \mathrm{GAP}(\mathbf{F}_{s,k}^{(i)}), \qquad
\mathbf{f}_{s,k}^{(i)} = \mathrm{normalize}\!\left(\mathbf{W}_q \bar{\mathbf{f}}_{s,k}^{(i)} + \mathbf{b}_q\right),
\end{equation}
where $\mathbf{W}_q \in \mathbb{R}^{d \times D_s}$ and $\mathbf{b}_q \in \mathbb{R}^{d}$ are shared query parameters and $\mathbf{F}_{s,k}^{(i)}$ denotes the synergy feature map produced by the multi-granularity interaction module for that sample.

\textbf{Hard Positive Mining.} We emphasize challenging within-task samples,
\begin{equation}
\mathcal{H}^{+}_k =
\left\{
i \in \mathcal{P}_k :
\cos(\mathbf{f}_{s,k}^{(i)}, \mathbf{p}_k)
< \cos(\theta_{\mathrm{hard}})
\right\},
\end{equation}
where $\mathbf{f}_{s,k}^{(i)}$ is the normalized synergy query defined above,
$\mathcal{P}_k$ indexes samples from task $k$, and
$\theta_{\mathrm{hard}} = 60^\circ$ is the hard-positive angular threshold.
This focuses the memory update on samples whose interaction patterns are
insufficiently aligned with the current task memory.

\textbf{Stochastic Memory Evolution.} Memory anchors are updated with momentum and controlled exploration noise:
\begin{align}
\mathbf{p}_k^{(t+1)} &= \text{normalize}\left(\gamma_{\text{proto}}\,\mathbf{p}_k^{(t)} + (1-\gamma_{\text{proto}})\,\boldsymbol{\mu}_k^{(t)} + \sigma_{\text{proto}}\,\boldsymbol{\xi}^{(t)}\right), \\
\boldsymbol{\mu}_k^{(t)} &= \frac{1}{|\mathcal{H}^{+}_k|} \sum_{i\in \mathcal{H}^{+}_k}\mathbf{f}_{s,k}^{(i)}, \quad \boldsymbol{\xi}^{(t)} \sim \mathcal{N}(\mathbf{0},\mathbf{I}),
\end{align}
with $\gamma_{\text{proto}} = 0.97$ and $\sigma_{\text{proto}} = 0.14$. If $|\mathcal{H}^{+}_k| = 0$, all samples in $\mathcal{P}_k$ are used.

\textbf{Memory Alignment Loss.} For each sample $i$ from task $k$, the same synergy query $\mathbf{f}_{s,k}^{(i)}$ is contrasted against all task memories $\{\mathbf{p}_j\}_{j=1}^K$:
\begin{equation}
\begin{split}
\mathcal{L}_{\text{align}} &= -\sum_{k=1}^{K} \sum_{i \in \mathcal{B}_k} \log \frac{\exp(\kappa\, \mathbf{f}_{s,k}^{(i)} \cdot \mathbf{p}_k)}{\sum_{j=1}^{K} \exp(\kappa\, \mathbf{f}_{s,k}^{(i)} \cdot \mathbf{p}_j)} \\
&\quad + \lambda_p \sum_{j\neq k}(\mathbf{p}_k^\top \mathbf{p}_j)^2,
\end{split}
\label{eq:L_align}
\end{equation}
where $\kappa = 16$, $\lambda_p = 0.01$, and the denominator simply measures how strongly the current sample's synergy query matches all memories, not just its own task anchor. Supplementary Sec.~S7 shows that hard positive updates remain more effective than hard negative mining in this setting.

\begin{figure}[t]
\centering
\includegraphics[width=\linewidth]{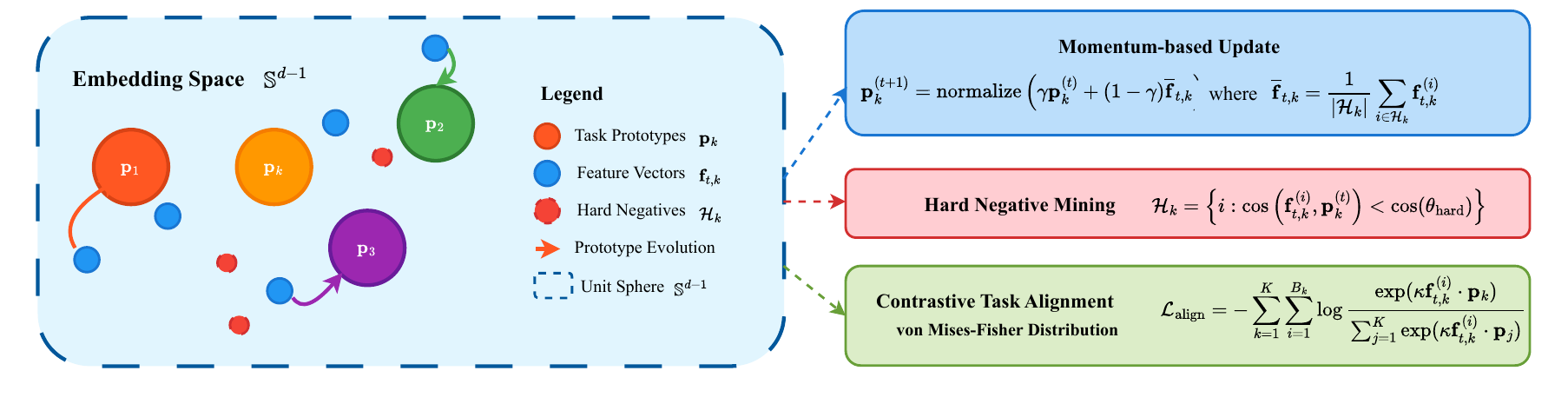}
\caption{Task-memory evolution with hard positive updates. Challenging within-task samples guide the memory anchor toward stable interaction patterns, improving synergy selection while keeping task anchors separated.}
\Description{
Illustration of task-memory evolution in a normalized embedding space.
Colored points denote feature representations from different tasks and
larger colored markers denote their corresponding task-memory anchors.
Samples that are poorly aligned with their own task anchor are identified
as challenging within-task examples and are emphasized during the memory
update. The selected features guide a momentum-based update of the
corresponding memory anchor, while the alignment objective attracts
features toward their own task memory and discourages different task
memories from collapsing together. The resulting task memories provide
task-dependent guidance for selecting cross-modal synergistic
interactions.
}
\label{fig:prototype_dynamics}
\end{figure}

\subsection{Multi-Granularity Cross-Modal Interaction for Synergy Composition}

Synergistic evidence is not assumed to exist as a static subspace. Instead, it is composed through cross-modal interaction at pixel, block, and global scales. We therefore keep the multi-granularity contrastive machinery, but interpret it as identifying interaction patterns that should contribute to $\mathbf{F}_s$.

\textbf{Pixel-level Interaction.} Let $\mathbf{z}_m^{(i,j)} \in \mathbb{R}^d$ denote the projected feature from modality $m \in \{1, 2\}$ at location $(i, j)$. The pixel-level objective is
\begin{equation}
\mathcal{L}_{\text{pixel}} = -\frac{1}{2HW}\sum_{i,j}\left[\log \frac{e^{\mathbf{z}_{1}^{(i,j)} \cdot \mathbf{z}_{2}^{(i,j)}/\tau_c}}{\sum_{i',j'}e^{\mathbf{z}_{1}^{(i,j)} \cdot \mathbf{z}_{2}^{(i',j')}/\tau_c}} + (1 \leftrightarrow 2)\right],
\end{equation}
which encourages fine-grained correspondence useful for constructing synergy.

\textbf{Block-level Interaction.} We divide features into non-overlapping blocks $\{\mathcal{B}_b\}_{b=1}^B$ and compute
\begin{equation}
\mathbf{z}_{b,m} = \sum_{(i,j) \in \mathcal{B}_b} \alpha_{i,j}^{(m)} \mathbf{h}_m^{(i,j)}, \quad \alpha_{i,j}^{(m)} = \frac{e^{\mathbf{w}^\top \mathbf{h}_m^{(i,j)}}}{\sum_{(i',j') \in \mathcal{B}_b} e^{\mathbf{w}^\top \mathbf{h}_m^{(i',j')}}},
\end{equation}
then apply the same InfoNCE form with contrastive temperature $\tau_c$ to $\{\mathbf{z}_{b,m}\}$ to capture locally compositional interactions.

\textbf{Global Interaction.} Global descriptors $\mathbf{g}_m = \text{GAP}(\mathbf{h}_m)$ are aligned via
\begin{equation}
\mathcal{L}_{\text{global}} = -\log \frac{e^{\mathbf{g}_1^\top \mathbf{g}_2 / \tau_c}}{\sum_{k=1}^{N} e^{\mathbf{g}_1^\top \mathbf{g}_k / \tau_c}} + \lambda_{\text{spectral}} \|\mathbf{g}_1 - \mathbf{g}_2\|_2^2,
\end{equation}
with $\lambda_{\text{spectral}} = 0.1$. This term stabilizes scene-level interaction statistics.

Granularity weights $w_g$ for $g \in \{\text{pixel}, \text{block}, \text{global}\}$ are computed adaptively via softmax over EMA-smoothed MI surrogates. We define the surrogate at granularity $g$ as
\begin{equation}
\widehat{\mathrm{MI}}_g^{(t)} =
\rho\, \widehat{\mathrm{MI}}_g^{(t-1)} +
(1-\rho)\left(-\mathcal{L}_g^{(t)} + \log |\mathcal{N}_g^{(t)}|\right),
\end{equation}
where $\mathcal{L}_g^{(t)}$ is the corresponding InfoNCE loss on batch $t$, $|\mathcal{N}_g^{(t)}|$ is the number of contrastive candidates at that granularity, and $\rho = 0.95$ is the EMA decay. The adaptive weights are then
\begin{equation}
w_g^{(t)} = \frac{\exp(\widehat{\mathrm{MI}}_g^{(t)} / \tau_w)}{\sum_{g'} \exp(\widehat{\mathrm{MI}}_{g'}^{(t)} / \tau_w)}.
\end{equation}
This quantity is a ranking surrogate rather than a calibrated MI estimator; it is used only to compare the relative interaction strength of the three granularities online. Supplementary Sec.~S6 gives the detailed implementation and ablation.

\subsection{Allocation Decoder and Training Objective}

The fused image is supervised by the standard reconstruction objective
\begin{align}
\mathcal{L}_{\text{fusion}} &= \lambda_{\text{int}} \underbrace{\| \mathbf{Y} - \max(\mathbf{I}_1, \mathbf{I}_2) \|_1}_{\mathcal{L}_{\text{intensity}}} \nonumber\\
&\quad + \lambda_{\text{grad}} \underbrace{\| \nabla \mathbf{Y} - \max(|\nabla \mathbf{I}_1|, |\nabla \mathbf{I}_2|) \|_1}_{\mathcal{L}_{\text{grad}}},
\label{eq:L_fusion}
\end{align}
where $\lambda_{\text{int}} = 1.0$ and $\lambda_{\text{grad}} = 10.0$.

We use Eq.~\eqref{eq:L_fusion} as a task-agnostic reconstruction prior rather than as a claim that all fusion domains obey the same physical image-formation model. The role of the max-intensity and gradient terms is simply to provide a common low-level supervision signal across benchmarks, while task-specific specialization is handled by the allocation streams and task memory. We therefore make the narrower claim that this shared objective is sufficient for controlled universal training on the evaluated benchmarks, not that it is the uniquely optimal loss for every fusion domain.

The decoder allocates information by combining shared structure with gated unique evidence and task-conditioned synergy:
\begin{equation}
\mathbf{s} = \sigma\big(\text{Conv}([\nabla \mathbf{I}_1, \nabla \mathbf{I}_2, \mathbf{F}_r])\big), \quad \widetilde{\mathbf{F}}_u = \mathbf{s} \odot [\mathbf{F}_{u1}, \mathbf{F}_{u2}],
\label{eq:gated_mix}
\end{equation}
where $\widetilde{\mathbf{F}}_u$ is the selectively preserved unique evidence. Synergy features $\mathbf{F}_s$ are produced by multi-granularity interaction conditioned on the current task memory, and the decoder input becomes $[\mathbf{F}_r, \widetilde{\mathbf{F}}_u, \mathbf{F}_s]$.

The unified objective is
\begin{equation}
\mathcal{L}_{\text{total}} = \mathcal{L}_{\text{fusion}} + \alpha \mathcal{L}_{\text{align}} + \beta \mathcal{L}_{\text{orthogonal}} + \gamma \sum_{g \in \{p, b, gl\}} w_g \mathcal{L}_g,
\label{eq:L_total}
\end{equation}
with default weights $\alpha = 0.5$, $\beta = 0.1$, and $\gamma = 1.0$. In this formulation, $\mathcal{L}_{\text{orthogonal}}$ protects non-interference, $\mathcal{L}_{\text{align}}$ stabilizes task memory, and the multi-granularity interaction terms determine which synergistic relations should be emphasized.

\begin{algorithm}[t]
\caption{RADIANT Training Procedure}
\label{alg:cmd_cta_training}
\begin{algorithmic}[1]
\REQUIRE Training data $\{(\mathbf{I}_1^{(n)}, \mathbf{I}_2^{(n)}, k^{(n)})\}_{n=1}^{N}$ with task labels $k^{(n)} \in \{1,\ldots,K\}$
\REQUIRE Learning rate $\eta$, initialized task memories $\{\mathbf{p}_k^{(0)}\}_{k=1}^{K}$, hyperparameters $(\alpha, \beta, \gamma, \gamma_{\text{proto}}, \sigma_{\text{proto}}, \kappa, \tau_c, \tau_w)$
\ENSURE Trained encoder $\phi_\theta$, decoder $\psi_\omega$, and task memories $\{\mathbf{p}_k^*\}_{k=1}^{K}$
\FOR{$t = 1$ to $T$}
\FOR{each mini-batch $\mathcal{B}_t$}
\STATE $\mathbf{F} \leftarrow \phi_\theta(\mathcal{B}_t)$
\STATE $\{\mathbf{F}_r, \mathbf{F}_{u1}, \mathbf{F}_{u2}\} \leftarrow \text{NonInterferingDecompose}(\mathbf{F})$
\STATE Compute $\mathcal{L}_{\text{orthogonal}}$ from pairwise overlap among $\mathbf{F}_r, \mathbf{F}_{u1}, \mathbf{F}_{u2}$
\FOR{each granularity $g \in \{\text{pixel}, \text{block}, \text{global}\}$}
\STATE Compute cross-modal interaction loss $\mathcal{L}_g^{(t)}$
\STATE Update EMA surrogate $\widehat{\mathrm{MI}}_g^{(t)} \leftarrow \rho\, \widehat{\mathrm{MI}}_g^{(t-1)} + (1-\rho)(-\mathcal{L}_g^{(t)} + \log |\mathcal{N}_g^{(t)}|)$
\ENDFOR
\STATE Compute adaptive weights $w_g^{(t)} \leftarrow \text{softmax}(\widehat{\mathrm{MI}}_g^{(t)} / \tau_w)$
\STATE Compose synergy features $\mathbf{F}_s$ through task-memory-guided multi-granularity interaction
\FOR{each task $k$}
\STATE Form pooled synergy queries $\mathbf{f}_{s,k}^{(i)}$ from $\mathbf{F}_{s,k}^{(i)}$
\STATE Mine hard positives $\mathcal{H}^{+}_k$
using $\theta_{\mathrm{hard}}$ and compute $\boldsymbol{\mu}_k^{(t)}$
\STATE Sample $\boldsymbol{\xi}^{(t)} \sim \mathcal{N}(\mathbf{0}, \mathbf{I})$ and update task memory $\mathbf{p}_k^{(t+1)}$
\ENDFOR
\STATE Compute $\mathcal{L}_{\text{align}}$ over task memories
\STATE $\mathbf{s} \leftarrow \sigma\big(\mathrm{Conv}([\nabla \mathbf{I}_1, \nabla \mathbf{I}_2, \mathbf{F}_r])\big)$
\STATE $\widetilde{\mathbf{F}}_u \leftarrow \mathbf{s} \odot [\mathbf{F}_{u1}, \mathbf{F}_{u2}]$
\STATE $\mathbf{Y} \leftarrow \psi_\omega([\mathbf{F}_r, \widetilde{\mathbf{F}}_u, \mathbf{F}_s])$
\STATE Compute $\mathcal{L}_{\text{fusion}}$ and $\mathcal{L}_{\text{total}} = \mathcal{L}_{\text{fusion}} + \alpha \mathcal{L}_{\text{align}} + \beta \mathcal{L}_{\text{orthogonal}} + \gamma \sum_g w_g \mathcal{L}_g$
\STATE Update $(\theta, \omega)$ via AdamW with gradient clipping
\ENDFOR
\ENDFOR
\end{algorithmic}
\end{algorithm}

\section{Experiments}

\begin{table*}[!tb]
\centering
\caption{Quantitative results on six fusion tasks. \textbf{Bold}: best; \underline{Underline}: second best. All metrics are higher-better. *: task-specific methods.}
\label{tab:main_results}
\resizebox{\textwidth}{!}{
\setlength{\tabcolsep}{3.2pt}
\begin{tabular}{lcccc|cccc|lcccc|lcccc}
\toprule
& \multicolumn{4}{c|}{(a1) MFIF: Lytro} & \multicolumn{4}{c|}{(a2) MFIF: MFI-WHU} & \multicolumn{5}{c|}{(c) MEIF: DSCIE} & \multicolumn{4}{c}{(d) NIR-VIS: Scene} \\
Method & EI & VIF & SCD & AG & EI & VIF & SCD & AG & Method & EI & VIF & SCD & AG & Method & EI & VIF & SCD & AG \\
\midrule
U2Fusion & 67.20 & 1.39 & 0.84 & 6.34 & 79.11 & 1.50 & 0.56 & 7.88 & U2Fusion & 83.00 & 1.69 & 0.49 & 8.71 & IFCNN & 82.13 & 0.90 & 1.14 & 8.72 \\
UNIFusion* & 70.14 & 1.30 & 0.60 & 6.77 & 66.57 & 1.01 & 0.29 & 7.19 & SPD-MEF* & 78.17 & 1.72 & 0.49 & 8.12 & U2Fusion & 80.73 & 1.07 & 1.19 & 8.51 \\
SDNet & 62.98 & 1.12 & 0.75 & 6.16 & 72.98 & 1.16 & 0.63 & 7.98 & MEF-GAN* & 80.21 & 1.59 & 0.63 & 8.06 & SDNet & 76.70 & 0.86 & 0.72 & 8.26 \\
MUFusion & 70.40 & 1.34 & 1.22 & 6.67 & 77.72 & 1.36 & 1.11 & 7.92 & MUFusion & 70.18 & 1.64 & 0.96 & 7.19 & MURF* & 41.71 & 0.41 & 0.19 & 4.22 \\
ZMFF* & 58.97 & 1.11 & 0.36 & 5.48 & 57.90 & 1.03 & 0.33 & 5.49 & IID-MEF* & 59.12 & 1.12 & 0.36 & 6.13 & Text-IF & 88.19 & 1.49 & 1.45 & 9.16 \\
\underline{GIF} & \underline{80.86} & \underline{1.74} & \underline{1.37} & \underline{7.71} & \underline{91.61} & \underline{1.94} & \underline{1.29} & \underline{9.25} & \underline{GIF} & \underline{111.27} & \underline{2.52} & \underline{1.04} & \underline{12.00} & \underline{GIF} & \underline{101.32} & \underline{1.51} & \underline{1.45} & \underline{10.83} \\
\textbf{RADIANT} & \textbf{85.55} & \textbf{1.86} & \textbf{1.45} & \textbf{8.16} & \textbf{97.93} & \textbf{2.08} & \textbf{1.38} & \textbf{9.89} & \textbf{RADIANT} & \textbf{118.73} & \textbf{2.71} & \textbf{1.12} & \textbf{12.88} & \textbf{RADIANT} & \textbf{108.20} & \textbf{1.62} & \textbf{1.56} & \textbf{11.62} \\
\midrule
& \multicolumn{4}{c|}{(b1) IVIF: LLVIP} & \multicolumn{4}{c|}{(b2) IVIF: TNO} & \multicolumn{5}{c|}{(e) Remote: QuickBird} & \multicolumn{4}{c}{(f) Medical: Harvard} \\
Method & EI & VIF & SCD & AG & EI & VIF & SCD & AG & Method & EI & VIF & SCD & AG & Method & EI & VIF & SCD & AG \\
\midrule
MURF & 38.02 & 0.25 & 1.56 & 3.75 & 45.93 & 0.94 & 1.52 & 4.35 & IFCNN & 18.30 & 1.16 & 0.86 & 1.73 & U2Fusion & 73.29 & 0.78 & 1.46 & 7.06 \\
LRRNet* & 34.93 & 0.34 & 0.95 & 3.64 & 36.37 & 0.75 & 1.40 & 3.75 & TextFusion & 13.73 & 0.76 & 0.32 & 1.29 & IFCNN & 98.67 & 0.92 & 1.33 & 9.57 \\
DDFM* & 41.13 & 0.51 & 1.55 & 4.43 & 31.01 & 0.67 & 1.60 & 3.03 & MUFusion & 18.45 & 1.05 & 0.89 & 1.72 & SDNet & 83.86 & 0.71 & 1.60 & 8.39 \\
CDDFuse* & 52.32 & 0.79 & 1.58 & 5.42 & 43.83 & 0.91 & 1.66 & 4.55 & P2Sharpen* & 16.66 & 1.20 & 0.94 & 1.56 & MUFusion & 88.66 & 0.97 & 1.23 & 8.39 \\
Text-IF* & 61.30 & 0.93 & 1.49 & 6.33 & 46.09 & 1.04 & 1.53 & 4.61 & ZeroSharpen* & 13.20 & 0.94 & 0.48 & 1.26 & CoCoNet* & 89.55 & 0.71 & 1.04 & 8.84 \\
\underline{GIF} & \underline{62.46} & \underline{0.73} & \underline{1.61} & \underline{6.70} & \underline{52.30} & \underline{0.99} & \underline{1.66} & \underline{5.24} & \underline{GIF} & \underline{23.02} & \underline{1.56} & \underline{1.04} & \underline{2.16} & \underline{GIF} & \underline{100.71} & \underline{1.10} & \underline{1.68} & \underline{9.73} \\
\textbf{RADIANT} & \textbf{66.08} & \textbf{0.78} & \textbf{1.70} & \textbf{7.09} & \textbf{55.33} & \textbf{1.06} & \textbf{1.78} & \textbf{5.62} & \textbf{RADIANT} & \textbf{24.85} & \textbf{1.67} & \textbf{1.12} & \textbf{2.32} & \textbf{RADIANT} & \textbf{107.06} & \textbf{1.18} & \textbf{1.80} & \textbf{10.44} \\
\bottomrule
\end{tabular}
}
\end{table*}

\subsection{Experimental Setup}

RADIANT is evaluated across six representative fusion tasks spanning diverse application domains: (1) \textbf{IVIF}: infrared-visible fusion on LLVIP~\citep{ref14} (12,025 pairs, 320 test) and TNO (40 pairs); (2) \textbf{MFIF}: multi-focus fusion on Lytro~\citep{ref28} (20 pairs) and MFI-WHU~\citep{ref46} (120 pairs); (3) \textbf{MEIF}: multi-exposure fusion on DSCIE (589 pairs); (4) \textbf{NIR-VIS}: near-infrared fusion on Scene (477 pairs); (5) \textbf{Remote sensing}: pansharpening on QuickBird (50 pairs); (6) \textbf{Medical imaging}: PET-MRI fusion on Harvard (302 pairs). Dataset splits follow official protocols; URLs and licensing details are provided in Supplementary Sec.~S0.

We employ five complementary metrics: Edge Intensity (EI) measuring edge preservation, Visual Information Fidelity (VIF)~\citep{ref26} quantifying information transfer, Sum of Correlation Difference (SCD) assessing structural consistency, Average Gradient (AG) evaluating detail richness, and Entropy (EN) measuring information content (reported in Supplementary Sec.~S4). Higher values indicate better performance for all metrics.

Inputs are resized to $256\times256$ and normalized per task using the standard preprocessing protocol of each benchmark. Across tasks, intensity normalization is kept modality-aware rather than globally shared so that modality-unique cues are not artificially collapsed before fusion.

Training uses AdamW optimizer (lr $2\times10^{-4}$, weight decay $5\times10^{-2}$) with cosine annealing, batch size 16, input resolution $256 \times 256$, and gradient clipping at 1.0. Loss weights are $\alpha = 0.5$, $\beta = 0.1$, $\gamma = 1.0$, and $\lambda_{\text{reg}} = 0.01$. Task-memory parameters are $\kappa = 16$, $\gamma_{\text{proto}} = 0.97$, $\sigma_{\text{proto}} = 0.14$, and $\lambda_p = 0.01$. The contrastive temperature is $\tau_c = 0.07$, the granularity-weight temperature is $\tau_w = 0.07$, and the hard-positive angular threshold is $\theta_{\mathrm{hard}} = 60^\circ$. Task memories are initialized from $k$-means centroids after one warm-up epoch. The tiny Vision Mamba--Swin backbone uses shallow dual branches with narrow channel widths (see Supplementary Sec.~S3), which explains the unusually small 3.108 MB footprint.

Systematic $\pm50\%$ perturbation sweeps across all hyperparameters yield $\leq 1.2\%$ average relative change in metrics (MAPD), demonstrating robust performance (see Supplementary Sec.~S1 for detailed sensitivity analysis). These results indicate limited sensitivity within the evaluated ranges.

We distinguish direct cross-dataset transfer from unlabeled test-time memory adaptation. In direct transfer, a model trained on LLVIP is evaluated directly on TNO without fine-tuning or target-memory updates. For test-time adaptation, all network weights remain frozen, while a new task memory is initialized from 50 unlabeled target samples and updated for 10 steps. Supplementary Sec.~S9 reports the corresponding sample-budget and update-step analyses.

\subsection{Main Results}

Table~\ref{tab:main_results} compares RADIANT with state-of-the-art methods across six fusion tasks. RADIANT consistently achieves the best performance, indicating that a single allocation strategy can adapt to heterogeneous fusion regimes without task-specific redesign. Results are averaged over 3 seeds; our significance statement refers specifically to comparisons against GIF, the strongest consistently available universal baseline across the six tasks.

\begin{figure*}[!htbp]
\centering
\includegraphics[width=0.95\textwidth]{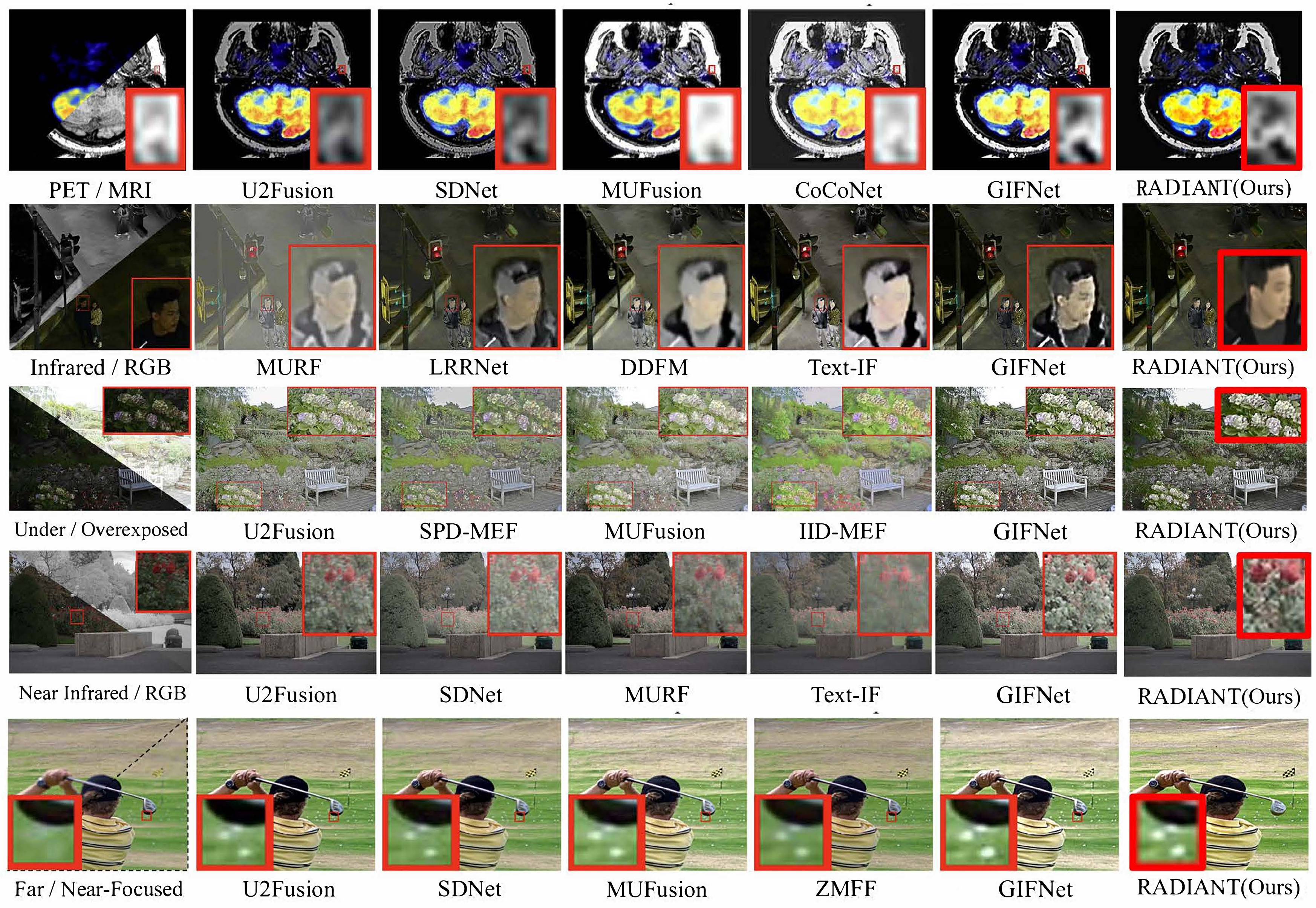}
\caption{Visual comparison across fusion tasks. Red boxes mark regions where allocation-aware fusion better preserves fine details and cross-modal complements.}
\Description{
Qualitative comparison of RADIANT with representative image-fusion
methods across five fusion scenarios. The rows show, from top to bottom,
PET--MRI medical fusion, infrared--RGB fusion, under- and over-exposed
image fusion, near-infrared--RGB fusion, and far- and near-focused image
fusion. Each row presents the source modalities followed by fused outputs
from several competing methods and RADIANT. Red boxes highlight the same
representative local region within each comparison. The highlighted
regions illustrate differences in anatomical and functional information
for PET--MRI fusion, target saliency and visible texture for
infrared--RGB fusion, local detail and luminance for exposure fusion,
scene texture for near-infrared--RGB fusion, and edge sharpness for
multi-focus fusion.
}
\label{fig:visual_comparison}
\end{figure*}

To evaluate whether the learned streams exhibit reduced overlap, Table~\ref{tab:allocation_diagnostics} reports two compact diagnostics on representative tasks: the reduction in shared--unique overlap $\Delta I(\mathbf{F}_r;\mathbf{F}_u)$ and the reduction in shared--synergy overlap $\Delta I(\mathbf{F}_r;\mathbf{F}_s)$, measured pre-/post-training using the estimators detailed in Supplementary Sec.~S2. The consistent negative shifts support the claim that RADIANT reduces harmful overlap while preserving task-relevant structure.

On FLIR thermal-visible detection using YOLOv5-L, RADIANT substantially outperforms GIF and single-modality baselines. Supplementary Sec.~S8 provides the complete detection protocol.

\begin{table}[!htbp]
\centering
\scriptsize
\caption{Compact information-allocation diagnostics on two representative tasks. Relative changes (\%) in KSG-estimated mutual information after training. Negative values indicate reduced overlap between the corresponding streams; negative $\Delta$ indicates reduced harmful overlap after training.}
\label{tab:allocation_diagnostics}
\begin{tabular}{lcc}
\toprule
Task & $\Delta I(\mathbf{F}_r;\mathbf{F}_u)$ & $\Delta I(\mathbf{F}_r;\mathbf{F}_s)$ \\
\midrule
LLVIP & $-23.2\%$ & $-19.7\%$ \\
Harvard & $-21.0\%$ & $-18.7\%$ \\
\bottomrule
\end{tabular}
\end{table}

\subsection{Efficiency Analysis}

Table~\ref{tab:efficiency} compares computational requirements. RADIANT maintains state-of-the-art fusion quality with only 3.108 MB parameters, showing that explicit information allocation need not incur the cost of heavyweight task-specific pipelines. The underlying backbone and projectors are intentionally shallow and narrow, and the extra allocation machinery is lightweight relative to competing universal models.

\begin{table}[!htbp]
\centering
\scriptsize
\caption{Computational efficiency comparison. Module-wise breakdown in Supplementary Table~S3.}
\label{tab:efficiency}
\begin{tabular}{l|l|ccc}
\toprule
Method & Technique & Size (MB) & GFLOPs & FPS \\
\midrule
DDFM & Diffusion model & 2211.00 & 1840.49 & 0.3 \\
Text-IF & ViLT & 580.81 & 82.85 & 8.2 \\
TC-MoA & MAE, MoE & 426.98 & 524.28 & 2.1 \\
GIF & Fusion only & 3.30 & 9.96 & 38 \\
\textbf{RADIANT} & \textbf{Allocation + Task Memory} & \textbf{3.108} & \textbf{9.42} & \textbf{42} \\
\bottomrule
\end{tabular}
\end{table}

The compact model size is achieved through depthwise separable convolutions in the encoder/decoder, weight-tied projectors for multi-granularity interaction, and shared lightweight MLP heads. The hybrid Vision Mamba--Swin backbone provides $O(N)$ complexity for long-range dependencies while maintaining local detail preservation.

\begin{table*}[!htbp]
\centering
\small
\caption{Ablation study on LLVIP. \textbf{Bold}: best results. NI: non-interfering shared/unique decomposition; Task memory: memory anchors with hard positive updates; Mamba/Swin: backbone branches.}
\label{tab:ablation}
\begin{tabular}{ccccc|cccc|c}
\toprule
NI & Task memory & Mamba & Swin & Multi-scale interaction & EI$\uparrow$ & VIF$\uparrow$ & SCD$\uparrow$ & AG$\uparrow$ & $\Delta$\% \\
\midrule
$\times$ & $\times$ & $\times$ & $\times$ & $\times$ & 52.31 & 0.61 & 1.42 & 5.18 & -- \\
$\checkmark$ & $\times$ & $\times$ & $\times$ & $\times$ & 60.12 & 0.70 & 1.55 & 6.30 & +14.9 \\
$\times$ & $\checkmark$ & $\times$ & $\times$ & $\times$ & 58.22 & 0.68 & 1.51 & 6.01 & +11.3 \\
$\checkmark$ & $\checkmark$ & $\times$ & $\checkmark$ & $\checkmark$ & 62.74 & 0.73 & 1.60 & 6.58 & +19.9 \\
$\checkmark$ & $\checkmark$ & $\checkmark$ & $\times$ & $\checkmark$ & 63.01 & 0.74 & 1.61 & 6.69 & +20.5 \\
$\checkmark$ & $\checkmark$ & $\checkmark$ & $\checkmark$ & $\checkmark$ & \textbf{66.08} & \textbf{0.78} & \textbf{1.70} & \textbf{7.09} & \textbf{+26.3} \\
\bottomrule
\end{tabular}
\end{table*}

\subsection{Ablation Study and Component Analysis}

Table~\ref{tab:ablation} evaluates the contribution of each component on LLVIP. The non-interfering decomposition and task memory both improve performance individually, and the strongest gains appear when they are paired with multi-scale interaction, supporting the view that universal fusion depends on coordinated allocation rather than on a single alignment block. The hybrid Mamba--Swin backbone further improves quality over either branch alone.

\subsection{Robustness and Generalization Analysis}

Table~\ref{tab:generalization} reports direct cross-dataset transfer from LLVIP to TNO without fine-tuning or target-time memory adaptation. RADIANT substantially outperforms GIF and CDDFuse, suggesting that the learned allocation policy transfers more faithfully across datasets than methods based on a single fusion regime.

\begin{table}[!htbp]
\centering
\scriptsize
\caption{Direct cross-dataset generalization (LLVIP $\rightarrow$ TNO) without fine-tuning or target-time adaptation. ``Full'' denotes training and evaluation on TNO, whereas ``Transfer'' denotes training on LLVIP and direct evaluation on TNO. Ratio = Transfer/Full $\times 100\%$.}
\label{tab:generalization}
\setlength{\tabcolsep}{4pt}
\begin{tabular}{l|cccc|cccc|c}
\toprule
& \multicolumn{4}{c|}{Full (trained on TNO)} & \multicolumn{4}{c|}{Transfer (LLVIP$\rightarrow$TNO)} & \\
Method & EI & VIF & SCD & AG & EI & VIF & SCD & AG & Ratio \\
\midrule
CDDFuse & 43.83 & 0.91 & 1.66 & 4.55 & 31.25 & 0.64 & 1.18 & 3.25 & 71.3\% \\
GIF & 52.30 & 0.99 & 1.66 & 5.24 & 40.17 & 0.76 & 1.27 & 4.02 & 76.8\% \\
\textbf{RADIANT} & \textbf{55.33} & \textbf{1.06} & \textbf{1.78} & \textbf{5.62} & \textbf{49.35} & \textbf{0.95} & \textbf{1.59} & \textbf{5.01} & \textbf{89.2\%} \\
\bottomrule
\end{tabular}
\end{table}

With 50 unlabeled TNO samples and 10 task-memory updates, EI increases from 49.35 to 50.16 while all network weights remain frozen. This setting is therefore unlabeled test-time memory adaptation rather than strict zero-shot inference; complete results are provided in Supplementary Sec.~S9.

Comparisons with cross-entropy, center loss, and static EMA memory are reported in Supplementary Sec.~S5.2. The complete task-memory formulation consistently achieves the best fusion quality and task accuracy.

\subsection{Information Allocation Validation}

Table~\ref{tab:subspace_isolation} evaluates how the learned streams contribute to reconstruction and detection. Although the allocation is only an approximation to true redundancy/uniqueness/synergy decomposition, the results show a functional division of labor: the shared stream preserves structure, the unique stream preserves modality-specific details, and the synergy-conditioned pathway contributes the strongest downstream discrimination.

\begin{table}[!htbp]
\centering
\small
\caption{Information allocation validation on LLVIP (reconstruction SSIM/PSNR) and FLIR (detection mAP@0.5). Metrics indicate complementary roles for shared, unique, and synergy-conditioned evidence.}
\label{tab:subspace_isolation}
\setlength{\tabcolsep}{8pt}
\begin{tabular}{l|cc|c}
\toprule
Stream Used & SSIM & PSNR & mAP@0.5 \\
\midrule
$\mathbf{F}_r$ only & 0.847 & 24.3 & 89.2\% \\
$[\mathbf{F}_{u1},\mathbf{F}_{u2}]$ only & 0.412 & 15.8 & 71.4\% \\
$\mathbf{F}_s$ only & 0.298 & 12.1 & 93.1\% \\
$\mathbf{F}_r + [\mathbf{F}_{u1},\mathbf{F}_{u2}]$ & 0.923 & 28.7 & -- \\
$\mathbf{F}_r + \mathbf{F}_s$ & -- & -- & 95.8\% \\
Full model & 0.941 & 30.2 & 96.8\% \\
\bottomrule
\end{tabular}
\end{table}

$\mathbf{F}_r$ preserves shared structure, the unique streams retain modality-specific details, and $\mathbf{F}_s$ captures interaction-dependent evidence that is especially beneficial for downstream tasks. Their combination yields the best overall performance.

Sensitivity to the hard-positive threshold is reported in
Supplementary Sec.~S7. Performance remains stable around the default
setting $\theta_{\mathrm{hard}}=60^\circ$.

\subsection{Visual Results and Analysis}

Figure~\ref{fig:visual_comparison} demonstrates RADIANT's visual performance across diverse fusion scenarios.

RADIANT demonstrates consistent visual superiority across diverse fusion tasks. In multi-focus scenarios, it preserves fine details without the blurring artifacts of competing methods. For infrared-visible fusion, it retains thermal saliency while preserving visible textures, reflecting better unique-evidence preservation. The method also produces more balanced luminance in exposure fusion and retains both anatomical and functional cues in PET/MRI fusion. These results support the central claim that universal fusion improves when redundant evidence is controlled, modality-unique evidence is preserved, and task-conditioned synergy is activated only where useful.

\section{Conclusion}

This paper presents RADIANT, a redundancy-aware information allocation framework for universal multi-modal fusion. RADIANT combines a non-interfering shared/unique parameterization with task-memory-guided synergy composition to allocate redundant, modality-unique, and task-conditioned synergistic information across fusion tasks. Experiments across six heterogeneous tasks show that this perspective yields strong universal performance, efficient inference, and robust transfer to unseen settings. More broadly, the results suggest that universal fusion benefits from controlling \emph{how} information is assigned within a shared model, not only from improving alignment strength or enlarging task-specific capacity. By combining a compact joint encoder with explicit shared/unique separation and memory-guided synergy composition, RADIANT provides a practical route to scalable multimodal fusion that remains lightweight, transferable, and interpretable across heterogeneous regimes.

\begin{acks}
This work was supported by the Research Institute of Newland Digital Technology.
\end{acks}

\bibliographystyle{ACM-Reference-Format}
\balance
\bibliography{sample-base}

\end{document}